\documentclass[journal,twoside,web]{ieeecolor}
\usepackage{generic}
\usepackage{amsmath,amssymb,amsfonts}
\usepackage{algorithm,algorithmic}
\usepackage{graphicx}
\usepackage{multirow}
\usepackage{textcomp}
\usepackage{array}
\usepackage{makecell}
\usepackage{soul}
\usepackage[export]{adjustbox} % 添加 adjustbox 宏包
\usepackage{bm}
\usepackage{tabularx}
\usepackage{booktabs} 
\usepackage{siunitx}
\usepackage{subcaption}
\makeatletter
\let\NAT@parse\undefined
\makeatother
\usepackage[colorlinks=true,
           linkcolor=blue,
           anchorcolor=blue,
           citecolor=green,
           urlcolor=blue,
           ]{hyperref}

\usepackage[numbers,sort&compress]{natbib}
\def\BibTeX{{\rm B\kern-.05em{\sc i\kern-.025em b}\kern-.08em
    T\kern-.1667em\lower.7ex\hbox{E}\kern-.125emX}}
\begin{document}
% \begin{CJK}{UTF8}{gbsn}

\title{Multimodal Feature Prototype Learning for Interpretable and Discriminative\\
Cancer Survival Prediction}

%Multimodal Feature Prototype Learning for Interpretable and Discriminative Cancer Survival Prediction

\author{Shuo Jiang, Zhuwen Chen, Liaoman Xu, Yanming Zhu, Changmiao Wang,\\ Jiong Zhang, Feiwei Qin, Yifei Chen, and Zhu Zhu
\thanks{This work was supported by the National Undergraduate Training Program for Innovation and Entrepreneurship under Grant 202510336053, the Fundamental Research Funds for
the Provincial Universities of Zhejiang under Grant GK259909299001-006, and the Anhui Provincial Joint Construction Key Laboratory of Intelligent Education Equipment and Technology under Grant IEET202401. \textit{(Corresponding authors: Zhu Zhu, Yifei Chen, and Feiwei Qin)}}
\thanks{Shuo Jiang, Zhuwen Chen, Liaoman Xu, and Feiwei Qin are with Hangzhou Dianzi University, Hangzhou 310018, China (e-mail: \{jiangshuo; 24050924; 23320302; qinfeiwei\}@hdu.edu.cn).}
\thanks{Yanming Zhu is with the School of Information and Communication Technology, Griffith University, QLD 4215, Australia (e-mail: yanming.zhu@griffith.edu.au).}
\thanks{Changmiao Wang is with Shenzhen Research Institute of Big Data, Shenzhen 518172, China (e-mail: cmwangalbert@gmail.com).}
\thanks{Jiong Zhang is with the Laboratory of Advanced Theranostic Materials and Technology, Ningbo Institute of Materials Technology and Engineering, Chinese Academy of Sciences, Ningbo 315201, China (e-mail: jiong.zhang@ieee.org).}
\thanks{Yifei Chen is with the School of Biomedical Engineering, Tsinghua University, Beijing 100084, China (e-mail: justlfc03@gmail.com).}
\thanks{Zhu Zhu is with Children's Hospital, Zhejiang University School of Medicine, National Clinical Research Center for Children and Adolescents' Health and Diseases, Hangzhou, 310052, China (e-mail: zhuzhu\_cs@zju.edu.cn).}
}

\maketitle

\begin{abstract}
Survival analysis plays a vital role in making clinical decisions. However, the models currently in use are often difficult to interpret, which reduces their usefulness in clinical settings. Prototype learning presents a potential solution, yet traditional methods focus on local similarities and static matching, neglecting the broader tumor context and lacking strong semantic alignment with genomic data. To overcome these issues, we introduce an innovative prototype-based multimodal framework, FeatProto, aimed at enhancing cancer survival prediction by addressing significant limitations in current prototype learning methodologies within pathology. Our framework establishes a unified feature prototype space that integrates both global and local features of whole slide images (WSI) with genomic profiles. This integration facilitates traceable and interpretable decision-making processes. Our approach includes three main innovations: (1) A robust phenotype representation that merges critical patches with global context, harmonized with genomic data to minimize local bias. (2) An Exponential Prototype Update Strategy (EMA ProtoUp) that sustains stable cross-modal associations and employs a wandering mechanism to adapt prototypes flexibly to tumor heterogeneity. (3) A hierarchical prototype matching scheme designed to capture global centrality, local typicality, and cohort-level trends, thereby refining prototype inference. Comprehensive evaluations on four publicly available cancer datasets indicate that our method surpasses current leading unimodal and multimodal survival prediction techniques in both accuracy and interpretability, providing a new perspective on prototype learning for critical medical applications. Our source code is available at \href{https://github.com/JSLiam94/FeatProto}{https://github.com/JSLiam94/FeatProto}.
\end{abstract}


\section{Introduction}
\label{sec:introduction}

\begin{figure}[tbp]
    \centering
    \includegraphics[width=\columnwidth]{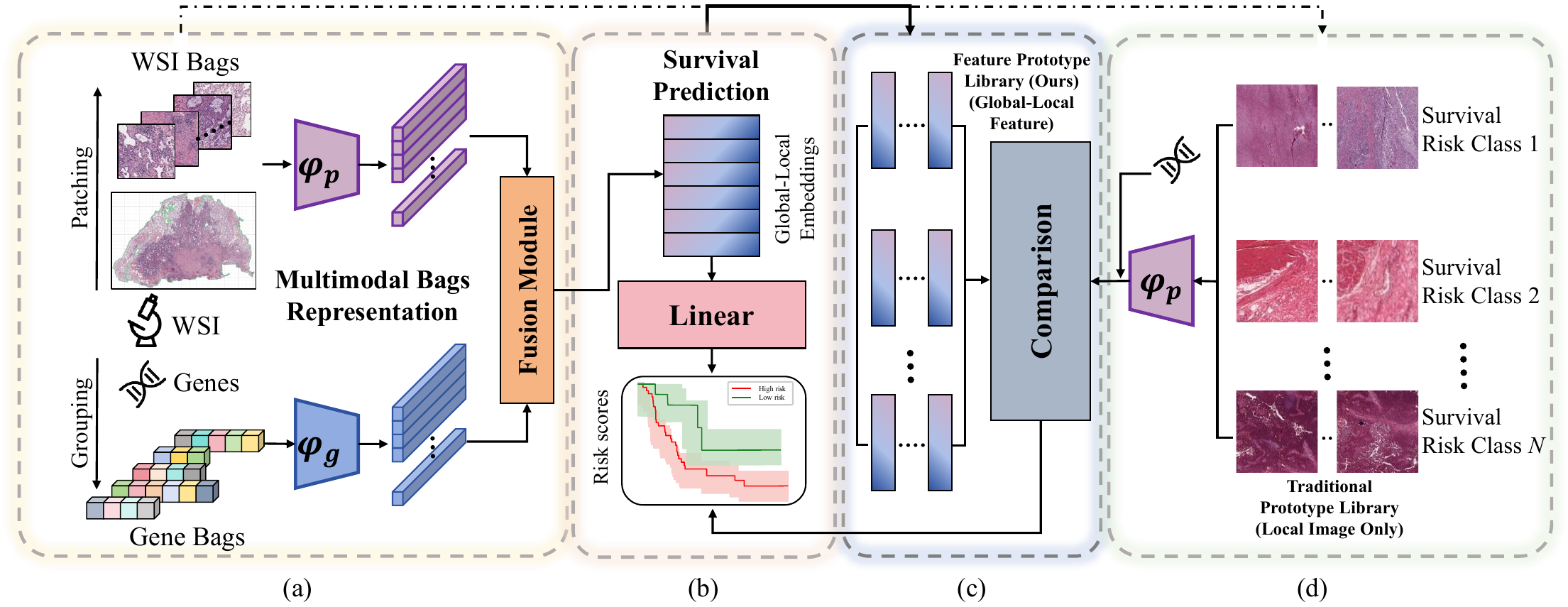} 
    \caption{Schematic of survival prediction. (a) Multimodal feature embedding \& fusion; (b) Conventional survival prediction (decision head: linear fully-connected layer); (c) Proposed feature prototype learning: prototype library with global-local fused features and multilevel deep prototype matching for accurate, interpretable survival prediction; (d) Traditional prototype library (local image only).}
    \label{fig: compare}
\end{figure}

\IEEEPARstart{S}{urvival} analysis plays a crucial role in cancer prognosis modeling, with its primary goal of assessing the probability of endpoint events (e.g., death) in specific patients over a future period and, accordingly, accurately stratifying patients' risk levels \cite{wiegrebe2024deep,fan2024stable}. This analysis not only reveals potential associations between disease progression and treatment responses, but also provides solid data support for formulating personalized treatment strategies that hold significant scientific research and clinical value \cite{zhang2024deep,monikapreethi2024survival}. In recent years, the rapid development of deep learning has driven the expansion of survival analysis from traditional clinical variables to a multimodal data paradigm \cite{zhou2024multimodal}. In particular, models that integrate pathological images from whole slide imaging (WSI) and genomic characteristics have become an essential direction for improving the precision of prognostic prediction \cite{zhou2025robust,chen2024accurate,xiong2024mome,li2024generalizable,chen2024camil}. The former provides visual clues regarding cellular morphology, tissue structure, and tumor heterogeneity, while the latter reveals biological mechanisms at the molecular level~\cite{wei2025annotation,zheng2024dynamic,Single11267275}. The two exhibit inherent and profound complementarity, facilitating synergistic and holistic modeling of cancer progression from both phenotypic traits and genotypic profiles \cite{ZHANG2026103521}.

Despite the initial performance achievements of existing multimodal survival analysis methods, most models still employ typical ``black-box" deep architectures, which lack sufficient interpretability, thereby limiting their application in clinical practice \cite{cui2025explainable, gou2025queryable,10494883Dual}. To address this issue, recent studies have emerged to improve model interpretability, including post-hoc explanation methods and intrinsically interpretable prototype learning methods \cite{WEN2025110901, 11039632,11464862,jia-etal-2026-scout}. The former performs post-hoc feature analysis on prediction results, while the latter introduces prototype learning during model training, aiming to bridge the gap between the model's latent feature space and human-understandable semantics \cite{wang2024mcpnet,dong2025disentangled}. However, traditional prototype learning methods typically rely on similarity analysis of local image regions, locating key regions through attention mechanisms, and performing visual matching with predefined prototype libraries \cite{ma2024interpretable, ayoobi2025protoargnet, wang2025mixture}. Such methods have particularly prominent limitations in the field of cancer pathology. Global features of the tumor microenvironment in WSIs are essential for prognosis \cite{zhu2025towards,nan2025deep}, but relying on single local information can introduce biases \cite{chen2022pan,chen2024sckansformer,11297150DGAN}. Additionally, the static nature and poor semantic alignment of prototype libraries can hinder the model's ability to identify key regions, impacting its discriminative power and interpretability \cite{chen2025mmlnb}.

To this end, this paper introduces an innovative approach to cancer survival analysis through a method that utilizes feature prototype learning, as illustrated in Fig. \ref{fig: compare}. The contributions of this study are threefold.
\begin{itemize}
    \item We develop a unified prototype space that effectively encodes both local and global weighted features from WSI and genomic profile representations. This integration facilitates precise and interpretable prognostic modeling, offering traceable pathways for decision-making and enhancing phenotypic characterization.
    \item We propose an Exponential Moving Average-based Prototype Update strategy (EMA ProtoUp). This strategy ensures the stability and adaptability of cross-modal associations. Additionally, our mechanism for Wandering Prototypes allows these prototypes to dynamically explore semantically relevant areas, which effectively addresses the challenge of tumor heterogeneity.

    \item We design a Multilevel deep Prototype Matching strategy (MPMatch). This strategy assesses cross-modal similarity from various perspectives, thereby enhancing prototype discrimination, expanding the diversity of the prototype library, and increasing the adaptability of the model.
\end{itemize}

\section{Related Work}

\subsection{Application of Multimodal Data Fusion in Cancer Survival Analysis}

As a critical component of precision medicine, cancer survival analysis has made remarkable progress in recent years, driven by the rapid advancement of deep learning \cite{jiang2024autosurv, ZHANG2025}. In particular, fusion of data from various modalities, such as WSI of pathological tissues and genomic data, enabled a more comprehensive characterization of tumor heterogeneity, thereby significantly improving prediction accuracy and clinical applicability. Genomic data, which provided molecular-level insights into cancer initiation and progression, have been widely used in tumor risk assessment and prognostic prediction \cite{10.1093/bib/bbad238, yu2025deep, 10.1093/bib/bbaf108}. On the other hand, pathological images, rich in visual information, have become a pivotal modality in cancer diagnosis and survival analysis \cite{yang2025foundation,yang2024scmil,9903546}. To fully exploit these complementary data sources, researchers have proposed various cross-modal fusion strategies. For instance, MOTCat \cite{xu2023multimodal} employed optimal transport theory to realize global alignment of cross-modal features. The cross-modal translation-alignment framework \cite{zhou2023cross} captured intra-modal information through parallel encoding-decoding structures and leveraged cross-modal attention to transfer complementary information. A memory-efficient multimodal Transformer \cite{jaume2023modeling} was used to fuse pathway and histological patch information, modeling inter-modal interactions for survival prediction. Multimodal knowledge decomposition (MKD) \cite{10669115} effectively separated redundant and modality-specific components, using cohort-guided modeling (CGM) to improve generalization performance by reducing noise. However, many multimodal models remain ``black-boxes", which complicates their use in clinical decision-making. Thus, effectively fusing multimodal information while simultaneously providing an intuitive and clinically meaningful explanatory mechanism remains a critical and pressing challenge in multimodal survival analysis.

\begin{figure*}[tbp] 
    \centering
    \includegraphics[width=2\columnwidth]{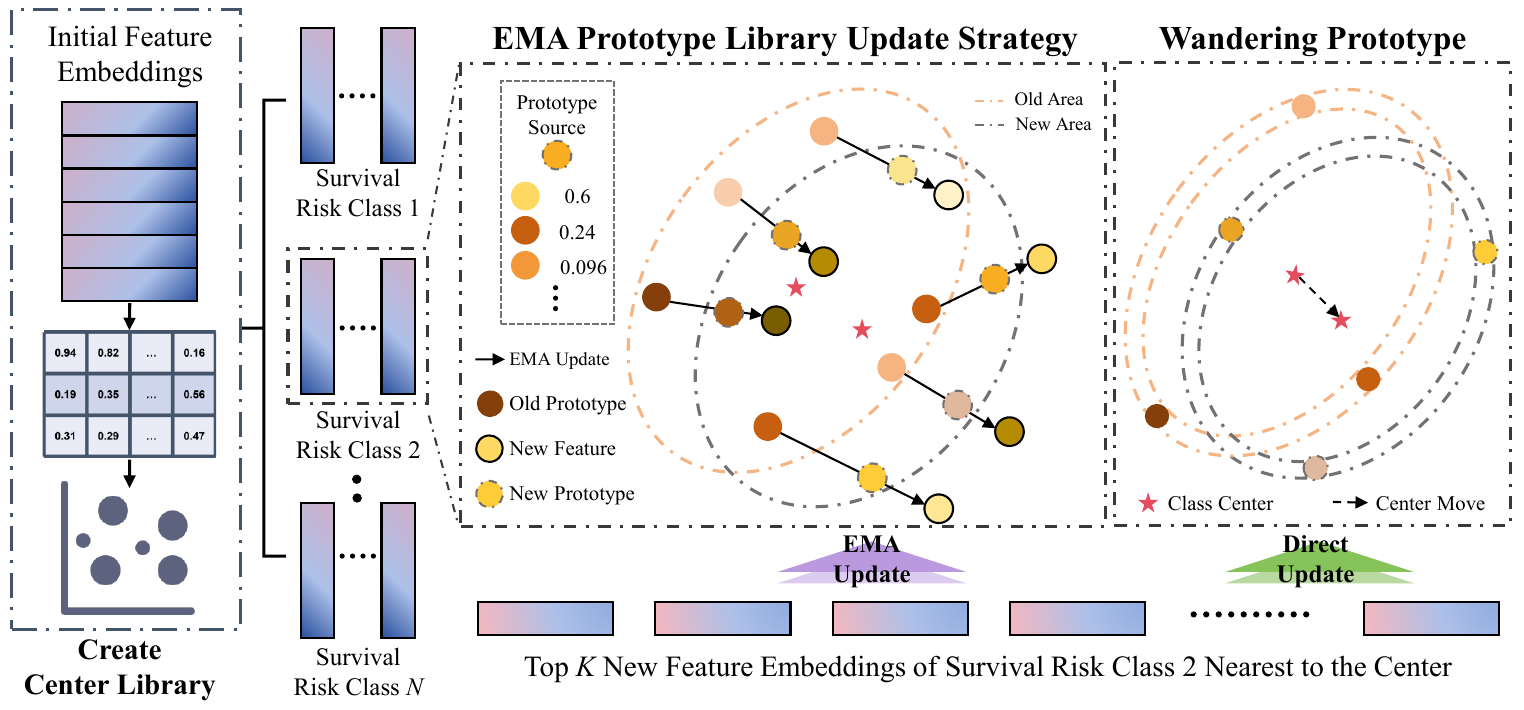} 
    \caption{Prototype Library Construction and Update. Left: Prototype library construction. The initial model generates feature embeddings to derive a similarity matrix, which is clustered to form a center library. Feature prototypes are organized by survival risk classes. Middle: EMA ProtoUp. Old prototypes migrate toward new features via EMA-based updates to generate new prototypes, representing typical samples. Right: Design and update mechanism of Wandering Prototypes. During prototype library construction or update, edge cases are selected from new features to represent special samples.}
    \label{fig: update_design}
\end{figure*}

\subsection{Research Progress in Explainable Methods and Prototype Learning}

To address the prevalent ``black-box" issue in deep learning models, explainable methods have garnered extensive attention. These methods are generally categorized into two types: post-hoc explanation and intrinsic explainability. Post-hoc explanation methods, such as Grad-CAM \cite{Selvaraju_2017_ICCV} and SHAP \cite{9265985}, analyzed the decisions of trained models and visualized regions of model focus through heatmaps. Intrinsic explainability methods enhance a model's interpretability by incorporating understandable structures during training, like prototype learning. This classic approach builds decision-making on ``explaining with examples." For instance, Prototypical Networks for few-shot learning \cite{snell2017prototypical} performed classification by calculating the similarity between input samples and prototypes. ProtoPNet \cite{chen2019looks} utilizes several typical prototype images. The model classifies images based on similarity comparisons with prototype images, enabling interpretable decision-making. Feature prototypes were introduced to represent discriminative features \cite{WEN2025110901}, letting models characterize their attributes across individuals and explain key image features for classification. While ProtoArgNet \cite{ayoobi2025protoargnet} and MGProto \cite{wang2025mixture} have advanced prototype learning, current methods primarily focus on classification through local similarity matching, limiting their effectiveness in survival analysis, which requires both global and local information.

\subsection{Exploration and Limitations of Prototype Learning in Medical Image Analysis}

Prototype learning holds enormous potential in the medical field, and preliminary studies have demonstrated that its explicit interpretive pathways can effectively boost clinical acceptance. For instance, PIP-Net \cite{nauta2023interpreting} enhanced the reliability and acceptability of the model by interpreting and correcting medical image classification results, while Proto-BagNets \cite{djoumessi2024actually} enabled transparent decision-making through joint local-global feature analysis. In the critical task of cancer survival prediction, prototype-based multimodal approaches have further advanced this direction: the MultiModal Prototyping (MMP) framework leverages morphological prototypes from WSIs and biological pathway prototypes from transcriptomic data for efficient multimodal fusion \cite{song2024multimodal}; The Prototypical Information Bottlenecking and Disentangling (PIBD) framework applies prototypical learning to select discriminative intra-modal instances and disentangle modality-common or specific knowledge, further optimizing multimodal cancer survival prediction \cite{zhang2023prototypical}. Despite these progresses, current mainstream cancer analysis methods still primarily concentrate on single imaging modalities and classification tasks, overlooking survival analysis that integrates imaging and genomic data \cite{gallee2023interpretable}. Traditional approaches often overlook global structural information and struggle to address cancer heterogeneity, thereby limiting their predictive performance. There remains an urgent need for a robust multimodal framework that seamlessly integrates both global and local features derived from WSIs and multi-omics genomic data: this integration not only enhances prediction accuracy and interpretability but also resolves complex associations related to tumor progression and patient prognosis.

\section{Method}

In this section, we present FeatProto in progressively increasing detail, from problem formulation to an implementable design. We first clarify the role of FeatProto and the discrete survival risk formulation in Section~\ref{sec:overall_architecture}. Sections~\ref{sec:prototype_design}, \ref{sec:prototype_init}, and~\ref{sec:prototype_update} then introduce the design, initialization, and dynamic update of the Feature Prototype Library. Section~\ref{sec:mpmatch} presents the multilevel prototype matching strategy for inference and explanation, and Section~\ref{sec:protosurv_loss} describes the ProtoSurv Loss for joint optimization of prototype discrimination and survival prediction.

\subsection{Overall Architecture}
\label{sec:overall_architecture}

Fig.~\ref{fig: update_design} illustrates the overall architecture of our proposed Feature Prototype Library (FeatProto), which explicitly models representative features for different survival risk classes. Our goal is to replace the conventional linear survival head with a prototype-based decision mechanism that enables traceable, interpretable, and discriminative risk prediction. Although cancer survival prediction is inherently a time-to-event task, our framework introduces discrete survival risk classes to support prototype construction and similarity-based reasoning. Specifically, following the commonly used discrete-time survival setting, we partition the continuous survival time axis into $C$ ordered intervals according to the distribution of uncensored training samples. Each interval corresponds to one survival risk class, and each patient is assigned a class label $y_i \in \{1,\dots,C\}$ based on the interval containing the observed survival time. Therefore, the term ``class'' in this paper refers to a discretized survival interval rather than a disease subtype or conventional semantic class.

Based on this formulation, we employ state-of-the-art (SOTA) survival prediction models and remove the fully connected linear head originally used for decision-making, using the remaining modules as the backbone for WSI-genomic feature extraction and fusion. FeatProto takes the final fused feature, immediately before the original linear survival head, as the input to construct and update the prototype library. The Feature Prototype Library stores $K_{\mathrm{typ}}$ representative features for each survival risk class as prototypes, each prototype is constructed from fused multimodal representations derived from WSI and genomic data. During inference, the model performs multi-strategy deep comparison by integrating nearest-prototype similarity, class-average similarity, and class-center similarity to estimate the survival risk class more accurately, thereby improving discriminative capability. Tracing the matched prototypes and their evolution further provides representative evidence for auxiliary diagnosis. Unlike traditional models that rely on a linear prediction head, this architecture substantially enhances decision interpretability.

This feature prototype library includes both typical and Wandering Prototypes to characterize the central tendency of each survival risk class as well as its intra-class diversity. It also supports dynamic updates to adapt to the continuously evolving feature space during training. The following sections describe its design, initialization, update mechanism, and integration with survival models in detail.

\subsection{Design of the Feature Prototype Library}
\label{sec:prototype_design}

The Feature Prototype serves as a ``feature anchor'' representing a specific survival risk class in the multimodal feature space. Here, each survival risk class corresponds to one discretized survival time interval. A feature prototype is explicitly defined as a feature vector that captures either the core characteristics (Typical Prototypes) or the marginal heterogeneity (Wandering Prototypes) of a given survival risk class within a unified space integrating global-local WSI features and genomic representations. Its core role is to establish a traceable decision-making path through ``input sample--prototype similarity matching,'' thereby linking model predictions to materialized multimodal evidence.

Accordingly, the feature prototype library is constructed in a class-specific manner, where each class corresponds to one discretized survival interval. It contains Typical Prototypes that capture the most representative patterns within that risk interval, and Wandering Prototypes that model diverse and atypical samples near the class boundary. This dual-prototype design balances representativeness and diversity, thereby improving robustness and generalization to unseen data.

\textbf{Typical Prototype.} For each survival risk class, the Typical Prototypes represent the most central features and encode the essential characteristics of patients belonging to that discretized survival interval.

\textbf{Wandering Prototype.} Wandering Prototypes are selected from regions moderately distant from the class center (within predefined boundaries) and are used to capture exceptional or boundary samples within the same survival risk class, thereby compensating for over-reliance on central features and improving robustness to edge cases.

The feature prototype library can be formalized as a tuple \( \mathcal{L}_{\mathrm{lib}} = (P, W, I, A) \), where:
\begin{itemize}
    \item \( P \in \mathbb{R}^{C \times K_{\mathrm{typ}} \times D} \) denotes Typical Prototypes ($C$: number of classes; $K_{\mathrm{typ}}$: number of prototypes per class; $D$: feature dimension);
    \item \( W \in \mathbb{R}^{C \times M_{\mathrm{wan}} \times D} \) denotes Wandering Prototypes ($M_{\mathrm{wan}}$: number of Wandering Prototypes per class);
    \item \( I \) is the mapping from prototype IDs to original data, including type and class index;
    \item \( A \) stores auxiliary information, such as mapping of associated data paths, for traceable interpretability analysis.
\end{itemize}

\subsection{Initialization of the Prototype Library}
\label{sec:prototype_init}

Prototypes are initialized from the training data in a data-driven manner. For each survival risk class $c \in \{1,\dots,C\}$, where $c$ denotes one discretized survival time interval defined in Section III-A, the extraction and fusion modules of the survival analysis backbone are used to integrate WSI and genomic information into a multimodal feature representation. We denote by
$
X_c = \{x_i \in \mathbb{R}^D \mid y_i = c\},
$
the set of fused features belonging to class $c$, where $y_i$ is the discretized survival risk label of sample $i$ obtained from the survival time binning procedure. The class center is calculated as follows:
\begin{equation}
\mu_c = \frac{1}{|X_c|} \sum_{x \in X_c} x.
\end{equation}
%我们提出幂平均距离相似度（Power Mean Distance Similarity）,通过先对每个距离  进行 m 次幂运算,可以显著放大特征之间的差异强调距离对于原型的重要性。
We propose the Power Mean Distance Similarity (PMDSim), where we increase each distance \(d_i\) to the $m$-th power, thus significantly amplifying the differences between features and emphasizing the importance of distance for prototypes:
\begin{equation}
S_p(a,b) = \frac{1}{1 + \left( \frac{1}{D} \sum_{i=1}^{D} |a_i - b_i|^m \right)},
\end{equation}
where $D$ denotes the feature dimension, \(|\cdot|\) represents the element-level absolute difference.

% 对于关键特征,幂运算可以增强这些特征在相似度计算中的权重,使得相似度指标更能反映病理学上的重要差异,而幂运算后的均值计算可以降低噪声或不相关的特征对相似度计算的干扰。通过调整幂次 m,可以更好地适应不同类型的特征,从而提高幂平均距离相似度计算的通用性和适应性。

Power operation enhances the weights of key features in the similarity calculation, better reflecting pathologically essential differences. Post-power mean calculation reduces noise/irrelevant feature interference. After L2 normalization of features \(x\) in \(X_c\), calculate the PMDSim with \(\mu_c\) and select the top $K_{\mathrm{typ}}$ features with the highest similarity for class \(c\).

%计算中L2归一化后的每个特征与的幂平均距离相似度,选择相似度最大的K个特征作为类别c的常规原型,代表类别的核心特征。
%对于游走原型,排除已选为常规原型的特征,计算剩余特征到的平均相似度。定义理想相似度范围(:游走边界),用于平衡类别内相似性与边缘特殊性。同时,确保c类别的游走原型不落在其他类别区域中。同样的,选取边界内与中心相似度最大的M个特征作为游走原型,代表类别的边缘特征与特殊案例。

For Wandering Prototypes, features already selected as Typical Prototypes are first excluded, and the average distance $\bar{d}_c$ from the remaining features to the class center $\mu_c$ is then computed. Based on this, we define an adaptive distance interval $[\bar{d}_c - \epsilon_c, \bar{d}_c + \epsilon_c]$ to better balance intra-class diversity and edge-case representativeness, while reducing the risk of overlap with other classes. Here, $\epsilon_c$ is defined as
$
\epsilon_c = \rho \bar{d}_c,
$
where $\rho$ is a learnable scaling parameter that dynamically adjusts the candidate range for selecting Wandering Prototypes during training. The top $M_{\mathrm{wan}}$ features within this interval are finally selected as Wandering Prototypes to capture informative edge cases and exceptional samples.

\subsection{EMA Prototype Library Update Strategy}
\label{sec:prototype_update}

The prototype library is updated every $n_{\mathrm{upd}}$ epochs. A straightforward update strategy is to directly replace outdated prototypes using newly observed representative features. Specifically, for each newly selected representative feature $x$ from survival risk class $c$, we compute its dissimilarity to each existing Typical Prototype $p_{c,i}$ in the same class:
\begin{equation}
d_{c,i} = 1 - S_p(x, p_{c,i}), \qquad i=1,\dots,K_{\mathrm{typ}}.
\end{equation}
A naive direct-replacement strategy would discard the least compatible prototype,
$
i_{\mathrm{far}} = \arg\max_i d_{c,i},
$
and replace it with the new feature $x$. However, such hard replacement causes abrupt shifts in the prototype space, which can harm training stability and generalization. On the other hand, simply retaining old prototypes may preserve outdated or unrepresentative features, resulting in prototype lag.

To address these issues, we propose an Exponential Moving Average Prototype Update strategy (EMA ProtoUp). Instead of directly replacing a prototype, the newly observed feature updates the most similar Typical Prototype in the same class through an exponential moving average. Let
$
i^\star = \arg\min_i d_{c,i}
$
denote the index of the closest Typical Prototype to $x$. The updated prototype is then computed as
\begin{equation}
p_{c,i^\star}^{\,\mathrm{new}}
=
\lambda p_{c,i^\star}^{\,\mathrm{old}}
+
(1-\lambda)x,
\qquad \lambda \in (0,1),
\end{equation}
where $\lambda$ controls the trade-off between historical prototype information and newly observed features.
This design allows the prototype library to evolve smoothly without abrupt structural changes. It improves adaptability to the evolving feature space while preserving stable class semantics during training.

For Wandering Prototypes, we do not apply EMA updates. Instead, at each prototype update step, they are re-selected from the latest candidate features according to the wandering-prototype selection rule described in Section~\ref{sec:prototype_init}. This strategy preserves boundary diversity and better captures atypical or edge-case samples.
In addition, we maintain a Prototype Source Library that stores the source sample ID, data path, and contribution weight for each prototype. During training, the top-$N_{\mathrm{src}}$ contributing source features are tracked for each updated prototype, which facilitates prototype tracing and improves interpretability.

\subsection{Multilevel Deep Prototype Matching Strategy}
\label{sec:mpmatch}
\begin{figure}[tbp] 
    \centering
    \includegraphics[width=\columnwidth]{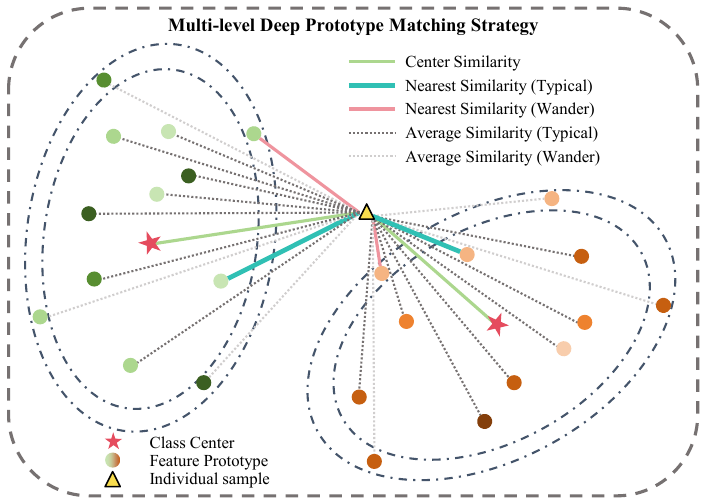} 
    \caption{Demonstration of MPMatch. By fusing center, class-average, and nearest-prototype similarities, it jointly optimizes local feature alignment and global class guidance, while improving interpretability through similarity visualization.}
    \label{fig: match}
\end{figure}

%多层次深度原型匹配策略
%传统原型学习仅仅依靠单一最近原型的比对策略容易受到异常样本的负面影响,造成生存预测准确度低下；同时仅仅依靠最近类别中心的比对策略在训练初期特征聚类不明显时存在显著劣势,容易带来模型欠拟合问题。为此,如图3所示,我们提出多层次深度原型匹配策略multilevel Deep Prototype Matching Strategy (MPMatch),通过特征在原型库不同层面的相似度融合,得出最优的生存预测决策。
Traditional prototype learning relies solely on the simplistic matching strategy of a single nearest prototype, which is highly vulnerable to adverse effects from outlier samples, resulting in notably low survival prediction accuracy. Meanwhile, relying solely on the matching strategy of the nearest class center has significant drawbacks when inter-class feature clustering is unclear in the early stages of training, which can easily lead to severe model underfitting. To address these issues, as shown in Fig. \ref{fig: match}, we propose the MPMatch. Through the fusion of similarities of features at different levels of the prototype library, including nearest-prototype responses and cohort-level trends captured by average and center similarities, optimal survival prediction decisions are derived.

%面对特征与原型库的深度匹配需求,通过多层次相似度融合与可解释性增强实现根据鲁棒性的决策。首先对输入特征张量完成设备适配与维度规整,针对每个类别动态构建有效特征原型集合,融合常规原型与游走原型,得到有效特征原型集合,每类共计个原型；
%对于输入特征与类别c的特征原型集合,基于幂平均距离相似度,度量输入特征与每一个特征原型的匹配度,并转换为相似度矩阵：
To achieve deep matching between features and the prototype library, decision-making relies on multilevel similarity fusion. First, device adaptation and dimension normalization are applied to the input feature tensor. For each class, an effective prototype set is constructed by fusing Typical Prototypes \(P_c\) and Wandering Prototypes \(W_c\), yielding $\mathcal{L}_{\mathrm{lib},c} = P_c \cup W_c$ with \(T_c\) prototypes. For the input feature \(f_b\) and class \(c\), based on PMDSim, the matching degree between \(f_b\) and each prototype \(p_{c,i}\) is measured and converted into a similarity matrix:
\begin{equation}
S_{b,c}
=
\big[
S_p(f_b, p_{c,1}),\dots,S_p( f_b, p_{c,T_c})
\big]
\in \mathbb{R}^{T_c}.
\end{equation}
%为了平衡局部原型匹配与全局类别引导,我们设计了三级相似度融合策略：
%类别平均相似度,计算特征原型集合的均值,反映特征与类别原型的整体匹配兼容性：
To balance local prototype matching and global class guidance, we design a three-level similarity fusion strategy:

\textbf{Class average similarity} measures the mean similarity between the input feature and all prototypes in class \(c\), reflecting the overall compatibility between the feature and that class.
\begin{equation}
    \bar S_{b,c} = \frac{1}{T_c} \sum_{i=1}^{T_c} S_p(f_b, p_{c,i}).
\end{equation}

%最近原型相似度,体现局部关键相应,提取原型集合中的最大相似度,聚焦最显著匹配的原型,实现局部特征对齐：
\textbf{Nearest prototype similarity} reflects local key responses, extracts the maximum similarity from the prototype set, focuses on the most prominently matched prototype, and achieves local feature alignment.
\begin{equation}
    S_{\mathrm{max},b,c} = \max_{i=1}^{T_c} S_p(f_b, p_{c,i}).
\end{equation}

%类别中心相似度,通过类别内特征原型的均值中心引入全局约束,引导决策的总体指向,避免局部原型偏差：
\textbf{Class center similarity} introduces global constraints through the mean center of intra-class feature prototypes, guides the overall direction of decision-making, and avoids local prototype bias.
\begin{equation}
  S_{\mathrm{center},b,c} = S_p(f_b, \mu_c).
\end{equation}

%通过预定义权重,加权融合以上三层次相似度,得到类别c对样本的最终生存预测决策:
By weighted fusion of the above three distinct levels of similarities using predefined weights \(\alpha_{\mathrm{sim}}\), \(\beta_{\mathrm{sim}}\), \(\gamma_{\mathrm{sim}}\), the final survival prediction decision for the sample \(f_b\) in class \(c\) is obtained as \(Logit_{b,c}\):
\begin{equation}
    Logit_{b,c} = \alpha_{\mathrm{sim}} \cdot \bar S_{b,c} + \beta_{\mathrm{sim}} \cdot S_{\mathrm{max},b,c} + \gamma_{\mathrm{sim}} \cdot S_{\mathrm{center},b,c}.
\end{equation}

%这一策略既平衡局部特征对齐与全局类别引导以提升决策鲁棒性,又可通过元信息展示决策依据,构建“特征匹配-原型溯源-临床解释”的可解释闭环,为生存分析场景下的原型库迭代与决策可解释性提供有力支撑。
Algorithm \ref{alg: algorithm_match} summarizes the inference-time prototype matching and explanation generation procedure of FeatProto. This strategy not only balances local feature alignment and global class guidance to improve decision-making robustness but also establishes an interpretable decision-making process based on prototype matching, source tracing, and clinical reasoning. This further supports prototype library iteration and enhances the interpretability of decisions in survival analysis scenarios.

For discrete-time survival prediction, the hazard probability at time bin $t$ is defined as
\begin{equation}
\lambda_t = \sigma(\mathrm{Logit}_t), \qquad \lambda_t \in (0,1).
\end{equation}
The survival probability up to time bin $t$ is computed as
\begin{equation}
S(t)=\prod_{k=1}^{t}(1-\lambda_k).
\end{equation}
The final scalar risk score used for ranking is defined as
\begin{equation}
R = -\sum_{t=1}^{T} S(t).
\end{equation}

% The risk score $h(s)$, which quantifies the likelihood of an event occurrence over time, is computed as:

% \begin{equation}
% h(s) = -\sum_{t} \left( \prod_{i=1}^{t} \left(1 - \sigma(Logit_i)\right) \right)
% \end{equation}

% where \(\sigma(\cdot)\) denotes the sigmoid activation function, transforming the model-generated logits (\(Logits_i\)) into hazard rates for each time point \(i\). The term \(\prod_{i=1}^{t} (1 - \sigma(Logit_i))\) represents the survival function at time \(t\), capturing the cumulative probability of event non-occurrence up to time \(t\) through the product of (1 - hazard rate) values across all preceding time points. The overall risk score is derived by summing these survival probabilities over all time points and taking the negative of the result, such that higher values correspond to an increased risk of event occurrence.

\begin{algorithm}[t]
\caption{Multilevel Deep Feature Prototype Library Matching with Explanation}
\label{alg: algorithm_match}
\begin{algorithmic}[1]
\REQUIRE Query features $\{q_b\}_{b=1}^{B}$, prototype library $\mathcal{L}_{\mathrm{lib}}$, source budget $N_{\mathrm{src}}$
\ENSURE Logits, similarity matrices, matched prototype IDs and types, and explanation dictionary
\STATE Initialize logits as a zero tensor of shape $[B, C]$
\STATE Initialize prototype origin storage for explanation
\FOR{each query feature $q_b$}
    \FOR{each class $c$}
        \STATE Compute similarities between $q_b$ and all prototypes in class $c$ using PMDSim
        \STATE Calculate $\bar{S}_{b,c}$, $S^{\max}_{b,c}$, and $S^{\mathrm{center}}_{b,c}$
        \STATE $Logit_{b,c} \leftarrow \alpha_{\mathrm{sim}} \bar{S}_{b,c} + \beta_{\mathrm{sim}} S^{\max}_{b,c} + \gamma_{\mathrm{sim}} S^{\mathrm{center}}_{b,c}$
        \STATE Find the nearest prototype and record its ID and type (\textit{Typical} or \textit{Wandering})
        \STATE Retrieve the top-$N_{\mathrm{src}}$ source samples of the matched prototype, together with their contribution weights
    \ENDFOR
\ENDFOR
\STATE \textbf{return} $Logit$, similarity matrices, prototype metadata, and explanation dictionary
\end{algorithmic}
\end{algorithm}

\begin{figure}[tbp] 
    \centering
    \includegraphics[width=\columnwidth]{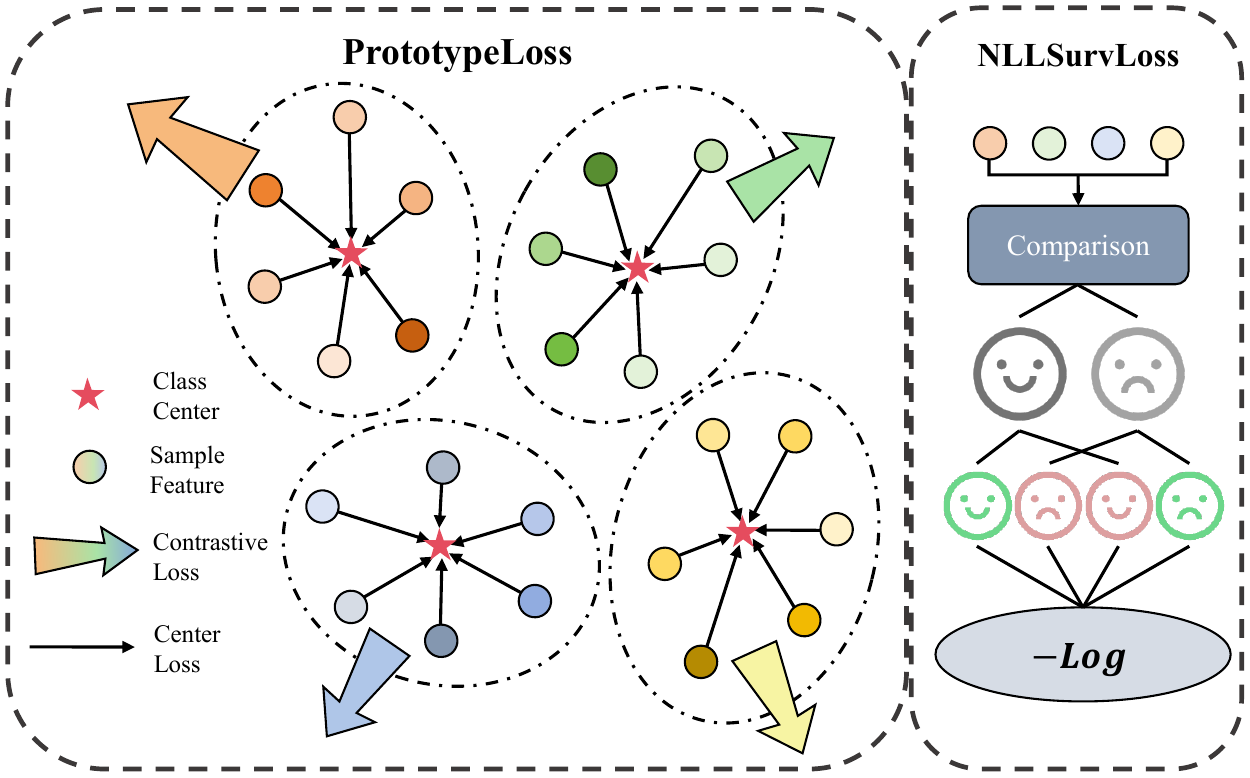} 
    \caption{Framework of ProtoSurv Loss for joint prototype-space optimization and survival supervision.}
    \label{fig:loss}
\end{figure}

\subsection{Prototype-Survival Fusion Loss}
\label{sec:protosurv_loss}

To jointly optimize prototype-space discrimination and survival prediction, we develop the Prototype-Survival Fusion Loss (ProtoSurv Loss), as shown in Fig.~\ref{fig:loss}.
Prototype Contrastive Loss increases feature similarity within the same class and enlarges the discrepancy between different classes, thereby promoting compactness and separability in the prototype space. For each sample feature \(f_i\), all effective prototypes of the same class, including both Typical Prototypes and Wandering Prototypes, are treated as positive prototypes \(P_i^{+}\), while all effective prototypes of other classes are treated as negative prototypes \(P_i^{-}\). The contrastive loss is defined as
\begin{equation}
\begin{gathered}
S_{ij}^{+}=S_p(f_i,p_j^{+}),  S_{ij}^{-}=S_p(f_i,p_j^{-}),\quad
p_j^{+}\in P_i^{+},  p_j^{-}\in P_i^{-},\\
L_{\mathrm{contra}}^{(i)}
=
-\log
\left(
\frac{
\sum_{p_j^{+}\in P_i^{+}} \exp(S_{ij}^{+})
}{
\sum_{p_j^{+}\in P_i^{+}} \exp(S_{ij}^{+})
+
\sum_{p_j^{-}\in P_i^{-}} \exp(S_{ij}^{-})
}
\right).
\end{gathered}
\end{equation}
Prototype Center Loss encourages sample features to stay close to the center of their corresponding class:
\begin{equation}
L_{\mathrm{center}}^{(i)} = 1 - S_p(f_i,\mu_{y_i}),
\end{equation}
where \(\mu_{y_i}\) denotes the prototype center of the class to which sample \(i\) belongs.

For survival prediction, we adopt the same discrete-time formulation introduced in Section~\ref{sec:mpmatch}. Let \(\lambda_t = \sigma(\mathrm{Logit}_t)\) denote the hazard probability at time bin \(t\), and let \(S(t)\) denote the corresponding survival probability.
Accordingly, the uncensored loss and censored loss for sample \(i\) are defined as
\begin{equation}
\begin{gathered}
L_{\mathrm{unc}}^{(i)}
=
-(1-c_i)\left[\log S(y_i-1)+\log \lambda_{y_i}\right],\\
L_{\mathrm{cen}}^{(i)}
=
-c_i \log S(y_i),
\end{gathered}
\end{equation}
where \(y_i\) denotes the discrete event-time bin of sample \(i\), and \(c_i\in\{0,1\}\) is the censoring indicator, with \(c_i=1\) for censored samples and \(c_i=0\) for uncensored samples.
The prototype loss is defined as
\begin{equation}
L_{\mathrm{proto}}
=
\frac{1}{B}\sum_{i=1}^{B}
\left(
L_{\mathrm{contra}}^{(i)}+\eta L_{\mathrm{center}}^{(i)}
\right),
\end{equation}
where \(\eta\) balances the contribution of the center loss.
The survival loss is defined as
\begin{equation}
L_{\mathrm{surv}}
=
\frac{1}{B}\sum_{i=1}^{B}
\left[
(1-\alpha_{\mathrm{loss}})L_{\mathrm{unc}}^{(i)}
+
\alpha_{\mathrm{loss}}L_{\mathrm{cen}}^{(i)}
\right].
\end{equation}
Finally, the total optimization objective is
\begin{equation}
L_{\mathrm{total}}
=
\beta_{\mathrm{loss}} L_{\mathrm{proto}}
+
(1-\beta_{\mathrm{loss}}) L_{\mathrm{surv}}.
\end{equation}
Together, the prototype library design, dynamic update mechanism, multilevel matching strategy, and ProtoSurv Loss form a complete prototype-based decision framework for interpretable multimodal survival prediction.

\section{Experiments}
\small
\begin{table*}[htbp]
\centering
\setlength{\tabcolsep}{6pt}
\begin{tabular}{l c c c c c}
\toprule
\multirow{2.5}{*}{\textbf{Model}} & \multicolumn{4}{c}{\textbf{Dataset}} & \multirow{2.5}{*}{\textbf{Overall}} \\
\cmidrule(lr){2-5}
& \textbf{LUAD (n = 453)}~\cite{albertina2016tcga} & \textbf{BLCA (n = 373)}~\cite{kirk2016tcga-blca} & \textbf{GBMLGG (n = 568)}~\cite{1882272492643518976,1884242817915227648} & \textbf{UCEC (n=480)}~\cite{erickson2016cancer} & \\ \hline
\multicolumn{6}{c}{\textbf{Genomics Only}} \\ \hline
MLP \cite{Taud2018} & 0.6046 $\pm$ 0.0417 & 0.6377 $\pm$ 0.0423 & 0.8228 $\pm$ 0.0194 & 0.6711 $\pm$ 0.0413 & 0.6841\\ 
SNN \cite{klambauer2017self} & 0.6198 $\pm$ 0.0340 & 0.6159 $\pm$ 0.0401 & 0.8312 $\pm$ 0.0251 & 0.6774 $\pm$ 0.0389 & 0.6861\\         
SNNTrans \cite{klambauer2017self, shao2021transmil} & 0.6379 $\pm$ 0.0451 & 0.6367 $\pm$ 0.0375 & 0.8237 $\pm$ 0.0206 & 0.6371 $\pm$ 0.0332 & 0.6839\\ \hline
\multicolumn{6}{c}{\textbf{WSI Only}} \\ \hline
AttenMIL \cite{ilse2018attention} & 0.6085 $\pm$ 0.0407 & 0.5614 $\pm$ 0.0537 & 0.7852 $\pm$ 0.0281 & 0.6532 $\pm$ 0.0386 & 0.6521\\ 
DeepAttnMISL \cite{yao2020whole} & 0.5610 $\pm$ 0.0519 & 0.5219 $\pm$ 0.0639 & 0.7660 $\pm$ 0.0337 & 0.6046 $\pm$ 0.0439 & 0.6134 \\ 
CLAM-SB \cite{lu2021data} & 0.5937 $\pm$ 0.0436 & 0.5600 $\pm$ 0.0583 & 0.7907 $\pm$ 0.0188 & 0.6517 $\pm$ 0.0318 & 0.6490 \\ 
CLAM-MB \cite{lu2021data} & 0.5893 $\pm$ 0.0571 & 0.5746 $\pm$ 0.0479 & 0.8036 $\pm$ 0.0245 & 0.6719 $\pm$ 0.0405 & 0.6621 \\ 
TransMIL \cite{shao2021transmil} & 0.6271 $\pm$ 0.0358 & 0.5693 $\pm$ 0.0421 & 0.7874 $\pm$ 0.0209 & 0.6710 $\pm$ 0.0437 & 0.6637 \\ \hline
\multicolumn{6}{c}{\textbf{Multimodal}} \\ \hline
MCAT \cite{Chen_2021_ICCV} & 0.6173 $\pm$ 0.0305 & 0.6675 $\pm$ 0.0352 & 0.8337 $\pm$ 0.0214 & 0.6503 $\pm$ 0.0418 & 0.6922\\ 
MOTCat \cite{xu2023multimodal} & 0.6746 $\pm$ 0.0327 & 0.6779 $\pm$ 0.0379 & 0.8417 $\pm$ 0.0199 & 0.6738 $\pm$ 0.0422 & 0.7170\\ 
Porpoise \cite{chen2022pan} & 0.6459 $\pm$ 0.0301 & 0.6437 $\pm$ 0.0403 & 0.8328 $\pm$ 0.0312 & 0.6867 $\pm$ 0.0320 & 0.7023\\ 
CMTA \cite{zhou2023cross} & 0.6763 $\pm$ 0.0348 & 0.6806 $\pm$ 0.0277 & 0.8476 $\pm$ 0.0233 & 0.6908 $\pm$ 0.0362 & 0.7238\\ 
SurvPath \cite{jaume2023modeling} & 0.6794 $\pm$ 0.0428 & 0.6743 $\pm$ 0.0341 & 0.8537 $\pm$ 0.0307 & 0.6815 $\pm$ 0.0351 & 0.7222\\ 
CCL \cite{10669115} & 0.6889 $\pm$ 0.0338 & 0.6810 $\pm$ 0.0289 & 0.8621 $\pm$ 0.0221 & 0.7038 $\pm$ 0.0369 & 0.7340\\ 
MMP \cite{song2024multimodal} & 0.6792 $\pm$ 0.0263 & 0.6804 $\pm$ 0.0279 & 0.8519 $\pm$ 0.0217 & 0.6916 $\pm$ 0.0332 & 0.7258\\ 
PIBD \cite{zhang2023prototypical} & 0.6746 $\pm$ 0.0334 & 0.6725 $\pm$ 0.0248 & 0.8441 $\pm$ 0.0195 & 0.6887 $\pm$ 0.0402 & 0.7200\\ 
MOTCat + FeatProto (Small) & 0.6884 $\pm$ 0.0314 & 0.6873 $\pm$ 0.0301 & 0.8514 $\pm$ 0.0203 & 0.6897 $\pm$ 0.0394 & 0.7292 \\ 
MOTCat + FeatProto (Large) & 0.6891 $\pm$ 0.0289 & 0.6804 $\pm$ 0.0364 & 0.8598 $\pm$ 0.0220 & 0.6990 $\pm$ 0.0421 & 0.7321 \\ 
CMTA + FeatProto (Small) & \underline{0.6972 $\pm$ 0.0378} & \underline{0.6887 $\pm$ 0.0324} & 0.8592 $\pm$ 0.0164 & 0.7049 $\pm$ 0.0391 & 0.7375 \\ 
CMTA + FeatProto (Large) & \textbf{0.7167 $\pm$ 0.0315} & 0.6755 $\pm$ 0.0398 & \underline{0.8757 $\pm$ 0.0186} & 0.7137 $\pm$ 0.0358 & \underline{0.7454}\\ 
CCL + FeatProto (Small) & 0.6963 $\pm$ 0.0271 & 0.6875 $\pm$ 0.0276 & 0.8739 $\pm$ 0.0191 & \underline{0.7189 $\pm$ 0.0384} & 0.7442 \\ 
CCL + FeatProto (Large) & 0.6947 $\pm$ 0.0302 & \textbf{0.6918 $\pm$ 0.0281} & \textbf{0.8774 $\pm$ 0.0152} & \textbf{0.7215 $\pm$ 0.0401} & \textbf{0.7464} \\ \bottomrule
\end{tabular}
\caption{C-index comparison across four datasets using ResNet-50 as the pathology feature extractor. The best and second-best scores are highlighted in \textbf{Bold} and \underline{Underline}, respectively. Overall represents the average C-index across datasets.}
\label{table: compare}
\end{table*}

\begin{table*}[htbp]
\centering
\setlength{\tabcolsep}{6pt}
\begin{tabular}{l c c c c c}
\toprule
\multirow{2.5}{*}{\textbf{Model}} & \multicolumn{4}{c}{\textbf{Dataset}} & \multirow{2.5}{*}{\textbf{Overall}} \\
\cmidrule(lr){2-5}
& \textbf{LUAD (n = 453)}~\cite{albertina2016tcga} & \textbf{BLCA (n = 373)}~\cite{kirk2016tcga-blca} & \textbf{GBMLGG (n = 568)}~\cite{1882272492643518976,1884242817915227648} & \textbf{UCEC (n=480)}~\cite{erickson2016cancer} & \\ \hline
MCAT~\cite{Chen_2021_ICCV} & 0.6191 $\pm$ 0.0368 & 0.6257 $\pm$ 0.0351 & 0.8109 $\pm$ 0.0273 & 0.6604 $\pm$ 0.0621 & 0.6790 \\ 
MOTCat \cite{xu2023multimodal} & 0.6097 $\pm$ 0.0321 & 0.6274 $\pm$ 0.0223 & 0.7949 $\pm$ 0.0371 & 0.6502 $\pm$ 0.0555 & 0.6706 \\ 
CMTA \cite{zhou2023cross} & 0.6651 $\pm$ 0.0605 & 0.6823 $\pm$ 0.0305 & 0.8527 $\pm$ 0.0213 & 0.7170 $\pm$ 0.0405 & 0.7293 \\ 
CCL \cite{10669115} & 0.6721 $\pm$ 0.0325 & 0.6796 $\pm$ 0.0314 & 0.8531 $\pm$ 0.0279 & \underline{0.7373 $\pm$ 0.0364} & 0.7355 \\ 
MOTCat + FeatProto (Large) & 0.6306 $\pm$ 0.0313 & 0.6662 $\pm$ 0.0276 & 0.8493 $\pm$ 0.0264 & 0.6875 $\pm$ 0.0428 & 0.7084 \\ 
CMTA + FeatProto (Large) & \underline{0.6757 $\pm$ 0.0276} & \textbf{0.6883 $\pm$ 0.0304} & \underline{0.8779 $\pm$ 0.0233} & 0.7204 $\pm$ 0.0371 & \underline{0.7406} \\ 
CCL + FeatProto (Large) & \textbf{0.6790 $\pm$ 0.0338} & \underline{0.6834 $\pm$ 0.0279} & \textbf{0.8782 $\pm$ 0.0240} & \textbf{0.7480 $\pm$ 0.0356} & \textbf{0.7472} \\ \bottomrule
\end{tabular}
\caption{C-index comparison across four datasets using CONCH as the pathology feature extractor, demonstrating the generalizability of FeatProto across different pathology encoders. Best results are in \textbf{bold}, and second-best are \underline{underlined}.}
\label{table: conch}
% \vspace{-10pt}
\end{table*}

\subsection{Experimental Setup}

\textbf{Dataset collection.} To demonstrate the performance of the proposed method, we utilized four publicly available cancer datasets from The Cancer Genome Atlas (TCGA): Lung Adenocarcinoma (LUAD, n = 453) \cite{albertina2016tcga}, Bladder Urothelial Carcinoma (BLCA, n = 373) \cite{kirk2016tcga-blca}, Glioblastoma \& Lower Grade Glioma (GBMLGG, n=568) \cite{1882272492643518976,1884242817915227648}, and Uterine Corpus Endometrial Carcinoma (UCEC, n=480) \cite{erickson2016cancer}. Each dataset contains paired diagnostic WSI and genomic data with basic survival outcomes. Genomic profiles provide molecular information about individuals, including RNA sequencing (RNA-seq), copy number variation (CNV), single-nucleotide variation (SNV), and DNA methylation. To conduct a fairer comparison, we follow existing research works \cite{zhou2023cross} and use RNA-seq, CNV, and SNV data, which were further divided into six subcategories: 1) tumor suppression, 2) tumorigenesis, 3) protein kinase, 4) cell differentiation, 5) transcription, and 6) cytokine and growth.

\begin{figure}[tbp]
    \centering
    \includegraphics[width=0.9\columnwidth]{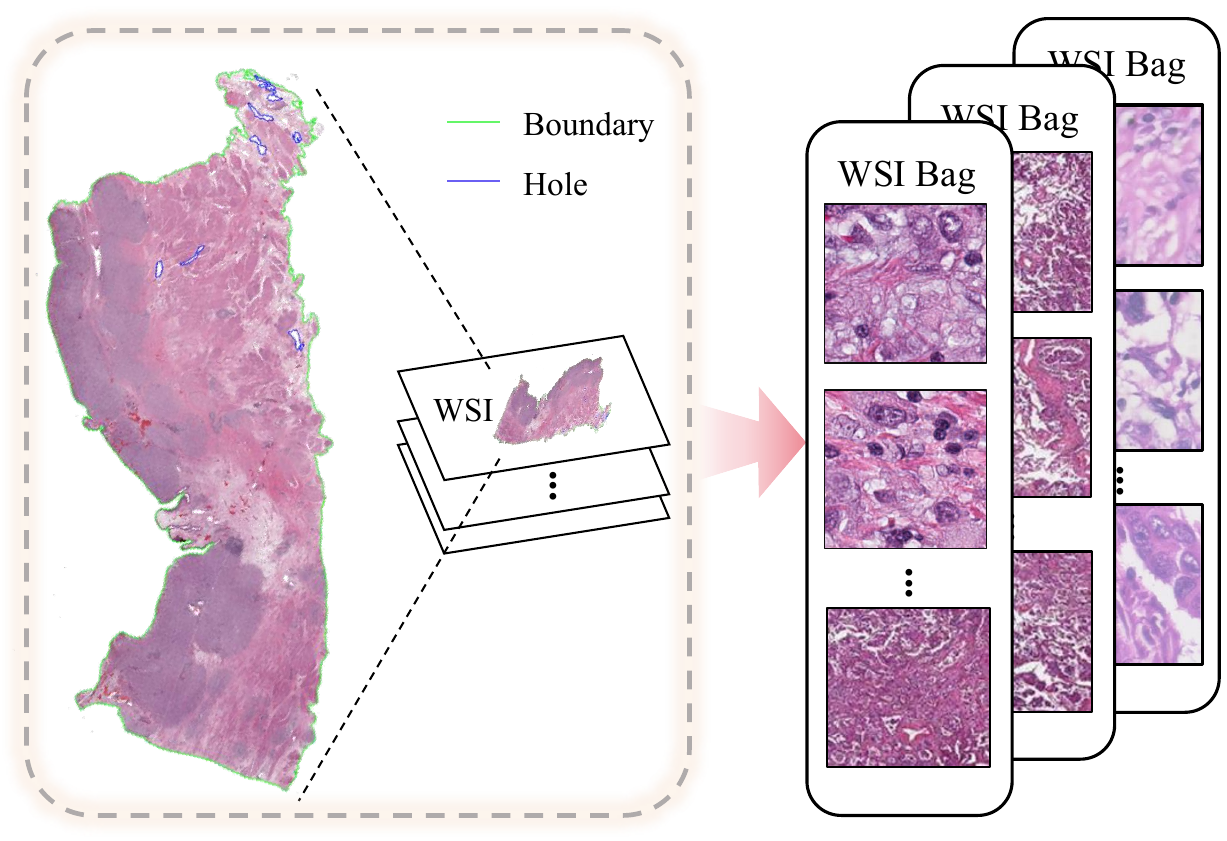} 
    \caption{The processing procedure of WSIs. After removing the edges and hole regions, the WSIs are segmented into patches, which are then combined to form WSI bags.}
    \label{fig: WSI}
\end{figure}

\begin{figure}[tbp]
    \centering
    \includegraphics[width=0.8\columnwidth]{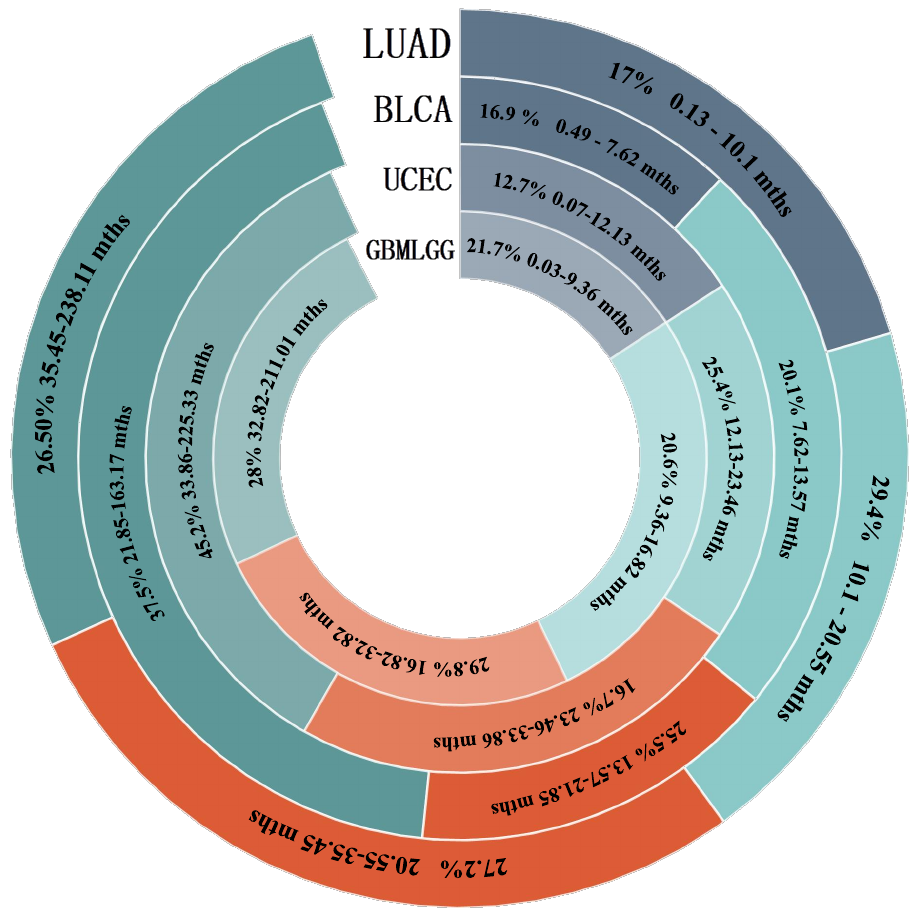} 
\caption{Visualization of survival time binning across all datasets. The time axis of each dataset is divided into four ordered survival intervals, which serve as survival risk classes for prototype construction and matching.}
    \label{fig: dataset}
\end{figure}

\textbf{Data preprocessing.} Following the general processing practices in current research, the CLAM \cite{lu2021data} model was employed for fully automated data preprocessing. For illustration, Fig. \ref{fig: WSI} depicts that for each WSI, automatic tissue segmentation was performed to exclude holes; subsequently, image patches capable of representing the entire region were extracted within these areas. The tissue regions were segmented into 256×256-sized patches at 20× magnification. For survival time binning, we adopt a discrete-time survival formulation and divide the survival time axis into $C=4$ ordered intervals according to the frequency distribution of uncensored training samples. The leftmost and rightmost intervals are further expanded to ensure that all patients can be assigned to one valid interval. Each interval is treated as a survival risk class and used as the class label for prototype construction and matching. Fig. \ref{fig: dataset} presents the temporal binning results of the four datasets.

% As illustrated in Fig. \ref{fig: PATCH}, t

% \begin{figure}[htbp]
%     \centering
%     \includegraphics[width=\columnwidth]{fig/image/patch.png} 
%     \caption{Architecture of the patch feature extraction model. A ResNet-50 \cite{he2016deep} network with residual connections processes input image patches, and the resulting patch-level features are generated for subsequent analysis tasks.}
%     \label{fig: PATCH}
% \end{figure} 

% \begin{table}[htbp]
% \small
% \centering
% \setlength{\tabcolsep}{10pt}
% \begin{tabular}{ c c }
% \hline
% \textbf{Model} & \textbf{Parameter Setting}\\
% \hline
% FeatProto (Large) & 40 Typical, 5 Wandering Prototypes \\ 
% FeatProto (Small) & 30 Typical, 5 Wandering Prototypes \\ 
% \hline
% \end{tabular}
% \caption{Prototype number settings for FeatProto.}
% \label{table: para}
% \end{table}

\textbf{Evaluation metrics.} The concordance index, known as the C-index, is a metric used to evaluate the performance of survival analysis models \cite{ALABDALLAH2024102781}. It quantifies the model's capability to correctly rank pairs of individuals according to their predicted survival outcomes. The C-index is formulated as follows:
\begin{equation}
\text{C-index} = \frac{1}{N} \sum_{i < j} \mathcal{I}(T_i < T_j) \cdot (1 - c_i) \cdot \mathcal{I}(R_i > R_j),
\end{equation}
where \(N\) denotes the total number of valid sample pairs; \(T_i\) and \(T_j\) represent the observed survival times of the \(i\)-th and \(j\)-th patients, respectively; \(R_i\) and \(R_j\) are the predicted risk scores defined in Section~\ref{sec:mpmatch}, with higher values indicating shorter expected survival times; \(\mathcal{I}(\cdot)\) is an indicator function; and \((1-c_i)\) ensures that the earlier individual is uncensored.

\textbf{Implementation details.} Experiments were performed on a system featuring a 32 GB NVIDIA Tesla V100-SXM2 GPU alongside an AMD EPYC 7542 32-Core Processor. The software setup included Python 3.8.20 and PyTorch 2.4.1, with CUDA 12.1 acceleration, and was deployed on Ubuntu 20.04.6 LTS. To guarantee reproducibility, a random seed of 2,026 was consistently used throughout all procedural operations. The AdamW optimizer \cite{loshchilov2017decoupled} was used with a learning rate of 1e-3 and cosine annealing learning-rate scheduling over 30 epochs. For each dataset, we adopted five-fold cross-validation to evaluate our model and other compared methods. For the primary comparison setting, all methods used an ImageNet-pretrained ResNet-50 \cite{he2016deep} for pathology feature extraction. Additional results using CONCH \cite{lu2024avisionlanguage} as the pathology feature extractor are reported in a separate table, demonstrating that FeatProto is not backbone-specific. The prototype update interval was set to $n_{\mathrm{upd}} = 1$ epoch. FeatProto (Large) was configured with 40 typical and 5 wandering prototypes, while FeatProto (Small) used 30 typical and 5 wandering prototypes.

%All methods used ResNet-50 \cite{he2016deep} to extract features when utilizing the pathological modality, and shared identical hyperparameters.

\subsection{Comparative Methods}

We evaluated our framework against various survival analysis models, including the genomics-based SNN \cite{klambauer2017self}, the pathology-based AttenMIL \cite{ilse2018attention}, and CLAM-SB \cite{lu2021data}, as well as other multimodal models. Below is an overview of selected multimodal competitors: MCAT \cite{Chen_2021_ICCV} uses a co-attention mechanism to integrate genomic and pathological features through separate Transformers for survival prediction. CMTA \cite{zhou2023cross} enhances modality-specific representations by integrating them with multimodal ones to convey complementary information. CCL \cite{10669115} breaks down multimodal knowledge, employing cohort-guided modeling to improve generalization.

As prototype learning methods, PIBD \cite{zhang2023prototypical} and MMP \cite{song2024multimodal} both employ prototypes for multimodal cancer survival prediction. PIBD learns risk-level prototypes to select discriminative intra-modal instances and disentangle multimodal data under a joint prototypical distribution. MMP summarizes gigapixel WSIs into morphological prototypes with Gaussian mixture models, encodes transcriptomic data into biological pathway prototypes, and fuses them for prediction.
Unlike these methods, our approach emphasizes decision interpretability by using a feature-prototype library to provide matching schemes and corresponding sample sources during risk stratification.

\begin{table}[tbp]

\centering
\setlength{\tabcolsep}{8pt}
\begin{tabular}{ l c }
\toprule
\textbf{Model} & \textbf{LUAD}~\cite{albertina2016tcga}\\
\hline
CMTA + FeatProto (Large) + Cosine Similarity & 0.6859$\pm$0.0293 \\ 
CMTA + FeatProto (Large) + Euclidean Similarity & 0.6941$\pm$0.0265 \\ 
CMTA + FeatProto (Large) + PMDSim & \textbf{0.7167$\pm$0.0315} \\ 
\hline
\textbf{Model} & \textbf{BLCA}~\cite{kirk2016tcga-blca}\\
\hline
CCL + FeatProto (Large) + Cosine Similarity & 0.6605$\pm$0.0277 \\ 
CCL + FeatProto (Large) + Euclidean Similarity & 0.6837$\pm$0.0302 \\ 
CCL + FeatProto (Large) + PMDSim & \textbf{0.6918$\pm$0.0281} \\ 
\bottomrule
\end{tabular}
\caption{Performance of different similarity metrics.}
\label{table: simi}
\end{table}

\begin{table}[htbp]
\centering
\setlength{\tabcolsep}{14pt}
\begin{tabular}{ c c c c}
% \hline
\toprule
\multirow{2.5}{*}{\textbf{$m$}} & \multicolumn{2}{c}{\textbf{Dataset}} & \multirow{2.5}{*}{\textbf{Overall}}\\
\cmidrule(lr){2-3} % 仅在Dataset标题下方绘制横线，覆盖LUAD和BLCA两列
 & \textbf{LUAD}~\cite{albertina2016tcga} & \textbf{BLCA}~\cite{kirk2016tcga-blca} & \\
\hline
1 & 0.6941$\pm$0.0265 & 0.6701$\pm$0.0267 & 0.6821\\
2 & \textbf{0.7167$\pm$0.0315} & \textbf{0.6755$\pm$0.0398} & \textbf{0.6961}\\
3 & 0.6770$\pm$0.0276 & 0.6431$\pm$0.0362 & 0.6601\\
4 & 0.6414$\pm$0.0352 & 0.6078$\pm$0.0344 & 0.6246\\
\bottomrule
\end{tabular}
\caption{Sensitivity to exponent $m$ in PMDSim of CMTA + FeatProto (Large).}
\label{table: m_of_sim}
\end{table}

\begin{table}[t]
\centering
\setlength{\tabcolsep}{10pt}
\begin{tabular}{ c c c c c }
\toprule
\multirow{2.5}{*}{\textbf{$\alpha_{\mathrm{sim}}$}}&\multirow{2.5}{*}{\textbf{$\beta_{\mathrm{sim}}$}} & \multirow{2.5}{*}{\textbf{$\gamma_{\mathrm{sim}}$}}&\multicolumn{2}{c}{\textbf{Dataset}}\\
\cmidrule(lr){4-5} % 仅在Dataset标题下方绘制横线，覆盖LUAD和BLCA两列
 &  &  & \textbf{LUAD}~\cite{albertina2016tcga} & \textbf{BLCA}~\cite{kirk2016tcga-blca}\\
\hline
0.33 & 0.33 & 0.33 & 0.6748$\pm$0.0269 & 0.6704$\pm$0.0272 \\
0.20 & 0.40 & 0.40 & 0.6759$\pm$0.0314 & 0.6653$\pm$0.0281 \\
0.40 & 0.20 & 0.40 & 0.6873$\pm$0.0297 & 0.6719$\pm$0.0339 \\
0.40 & 0.40 & 0.20 & \textbf{0.7167$\pm$0.0315} & \textbf{0.6755$\pm$0.0398} \\
\bottomrule
\end{tabular}
\caption{Results of different parameters for the three-level similarity of CMTA + FeatProto (Large) within MPMatch.}
\label{table: set}
\end{table}

\begin{table}[t]
\centering
\setlength{\tabcolsep}{13pt}
\begin{tabular}{ c c c c}
\toprule
\multirow{2.5}{*}{\textbf{$\lambda$}} & \multicolumn{2}{c}{\textbf{Dataset}} & \multirow{2.5}{*}{\textbf{Overall}}\\
\cmidrule(lr){2-3} % 仅在Dataset标题下方绘制横线，覆盖LUAD和BLCA两列
 & \textbf{LUAD}~\cite{albertina2016tcga} & \textbf{BLCA}~\cite{kirk2016tcga-blca} & \\
\hline
0.01 & 0.6837$\pm$0.0263 & 0.6615$\pm$0.0372 & 0.6726\\
0.05 & 0.6959$\pm$0.0330 & 0.6691$\pm$0.0343 & 0.6825\\
0.10 & \textbf{0.7167$\pm$0.0315} & \textbf{0.6755$\pm$0.0398} & \textbf{0.6961}\\
0.15 & 0.6806$\pm$0.0274 & 0.6714$\pm$0.0387 & 0.6760\\
\bottomrule
\end{tabular}
\caption{Hyperparameter experiment of $\lambda$ in EMA ProtoUp on CMTA + FeatProto (Large).}
\label{table: lambda}
\end{table}

\subsection{Quantitative Evaluation}

As shown in Table \ref{table: compare}, when we replaced the original fully connected linear layers of three SOTA deep learning architectures (MOTCat, CMTA, and CCL) with our model, we observed a significant improvement in their C-index scores for traceable and interpretable survival prediction. Specifically, on the LUAD dataset, our modifications resulted in C-index scores of 68.91\%, 71.67\%, and 69.63\%, representing improvements of 1.45\%, 4.04\%, and 0.74\% compared to the original models, respectively. Similar performance improvements were observed on the BLCA dataset. The improvement on the GBMLGG dataset was also notable, with a maximum increase of 2.81\%, while the most significant improvement was observed on the UCEC dataset, with an average increase of 2.19\%. 

These results indicate that our framework outperforms other advanced multimodal survival models while providing a traceable and interpretable decision-making process. Notably, 40 prototypes performed better on LUAD, GBMLGG, and UCEC, whereas 30 prototypes performed better on BLCA. This differs from the traditional view that favors few-shot learning over conventional prototype learning. In contrast, our framework benefits from multiple feature prototypes, especially on smaller datasets affected by outliers.
Furthermore, compared to single-modal models, our framework outperformed the previous best models by 7.88\%, 5.41\%, 4.62\%, and 4.41\% on the four datasets, respectively, indicating significant improvements.

\begin{figure*}[htbp] 
    \centering
    \includegraphics[width=2\columnwidth]{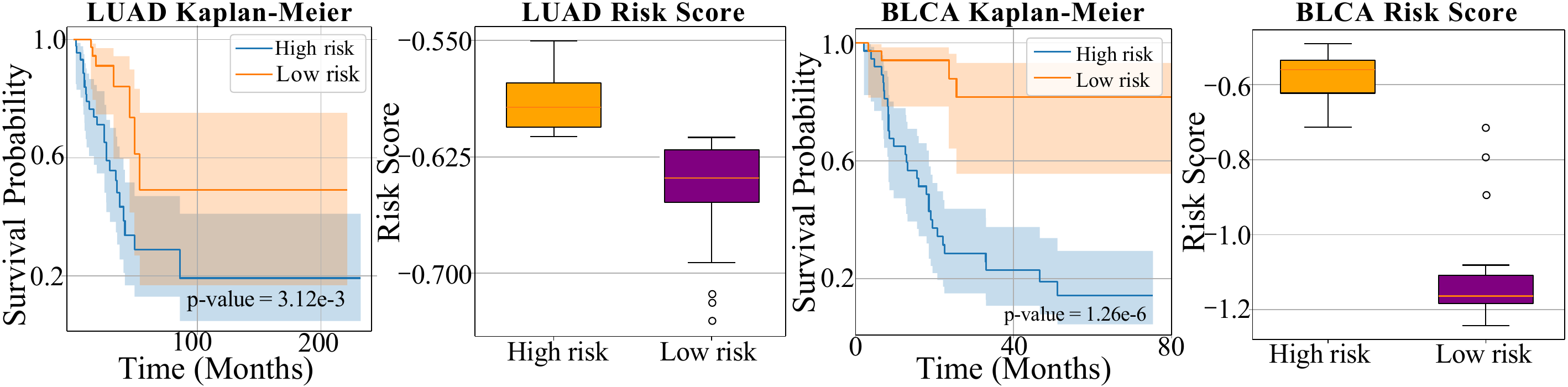} 
    \caption{Kaplan-Meier survival curves and box plots of risk score distribution on the LUAD and BLCA datasets.}
    \label{fig: kaplan}
\end{figure*}

Kaplan-Meier analysis is a nonparametric statistical method used to estimate the survival function based on survival time data. Specifically, we divided patients into high- and low-risk groups based on the cohort's median predicted risk score, represented by the red and blue curves, respectively. In Kaplan-Meier analysis, the p-value represents the probability of observing differences in survival rates between the two groups under the null hypothesis that there is no actual difference between them. As shown in Fig. \ref{fig: kaplan}, the p-values obtained by our method are significantly lower than 0.05 in both the LUAD and BLCA datasets, indicating a statistically significant distinction between the high-risk and low-risk groups.

To evaluate backbone generalizability, Table \ref{table: conch} reports results using CONCH as the pathology feature extractor. The results show that FeatProto remains effective under this alternative encoder and continues to improve the corresponding baselines, supporting its compatibility with different pathology feature extractors.

\begin{table}[tbp]
\centering
\setlength{\tabcolsep}{9pt}
\begin{tabular}{ l c c c}
\toprule
\multirow{2.5}{*}{\textbf{Loss}} & \multicolumn{2}{c}{\textbf{Dataset}} & \multirow{2.5}{*}{\textbf{Overall}}\\
\cmidrule(lr){2-3} % 仅在Dataset标题下方绘制横线，覆盖LUAD和BLCA两列
 & \textbf{LUAD}~\cite{albertina2016tcga} & \textbf{BLCA}~\cite{kirk2016tcga-blca} &\\
\hline
NLLSurvLoss & \underline{0.6779$\pm$0.0337} & \underline{0.6579$\pm$0.0371} & \underline{0.6679}\\
ProtoLoss & 0.6371$\pm$0.0381 & 0.6318$\pm$0.0391 & 0.6345\\
ProtoSurv Loss & \textbf{0.7167$\pm$0.0315} & \textbf{0.6755$\pm$0.0398} & \textbf{0.6961}\\
\bottomrule
\end{tabular}
\caption{Performance of CMTA + FeatProto (Large) under different loss functions.}
\label{table: loss_1}
\end{table}

% \begin{figure}[tbp]
%     \centering
%     \includegraphics[width=\columnwidth]{fig/image/UMAP.png} 
%     \caption{UMAP Dimensionality Reduction Plots of Feature Prototype Library on LUAD and BLCA datasets.}
%     \label{fig: UMAP}
% \end{figure} 

\begin{table}[t]
\centering
\small
\setlength{\tabcolsep}{4.4pt}
\begin{tabular}{l c c c}
\toprule
% 新增Dataset标题行，下方通过\cline{2-3}添加仅覆盖两数据集列的横线
\multirow{2.5}{*}{\textbf{Model}} & \multicolumn{2}{c}{\textbf{Dataset}} & \multirow{2.5}{*}{\textbf{Overall}}\\
\cmidrule(lr){2-3} % 仅在Dataset标题下方绘制横线，覆盖LUAD和BLCA两列
& \textbf{LUAD}~\cite{albertina2016tcga} & \textbf{BLCA}~\cite{kirk2016tcga-blca} &  \\
\hline
CMTA + FeatProto (Large) & \textbf{0.7167} & \textbf{0.6755} & \textbf{0.6961} \\
w/o Wandering Prototype & \underline{0.6903} & \underline{0.6703} & \underline{0.6803}\\ 
w/o EMA ProtoUp & 0.6798 & 0.6608 & 0.6703\\ 
w/o MPMatch & 0.6670 & 0.6641 & 0.6656\\ 
% \hline
\toprule
% 第二部分保持与第一部分一致的格式，Dataset下方同样添加横线
\multirow{2.5}{*}{\textbf{Model}} & \multicolumn{2}{c}{\textbf{Dataset}} & \multirow{2.5}{*}{\textbf{Overall}}\\
\cmidrule(lr){2-3} % 仅在Dataset标题下方绘制横线，覆盖LUAD和BLCA两列
& \textbf{LUAD}~\cite{albertina2016tcga} & \textbf{BLCA}~\cite{kirk2016tcga-blca} &  \\
\hline
CCL + FeatProto (Large)  & \textbf{0.6947} & \textbf{0.6918} & \textbf{0.6932}\\ 
w/o Wandering Prototype & \underline{0.6911} & \underline{0.6889} & \underline{0.6900}\\ 
w/o EMA ProtoUp  & 0.6848 & 0.6845 & 0.6846\\ 
w/o MPMatch  & 0.6779 & 0.6728 & 0.6754\\ 
\bottomrule
\end{tabular}
\caption{Ablation results on LUAD and BLCA in terms of C-index, validating the effectiveness of each component.}
\label{table: ablation}
\end{table}

\begin{figure*}[tbp]
    \centering
    \includegraphics[width=2\columnwidth]{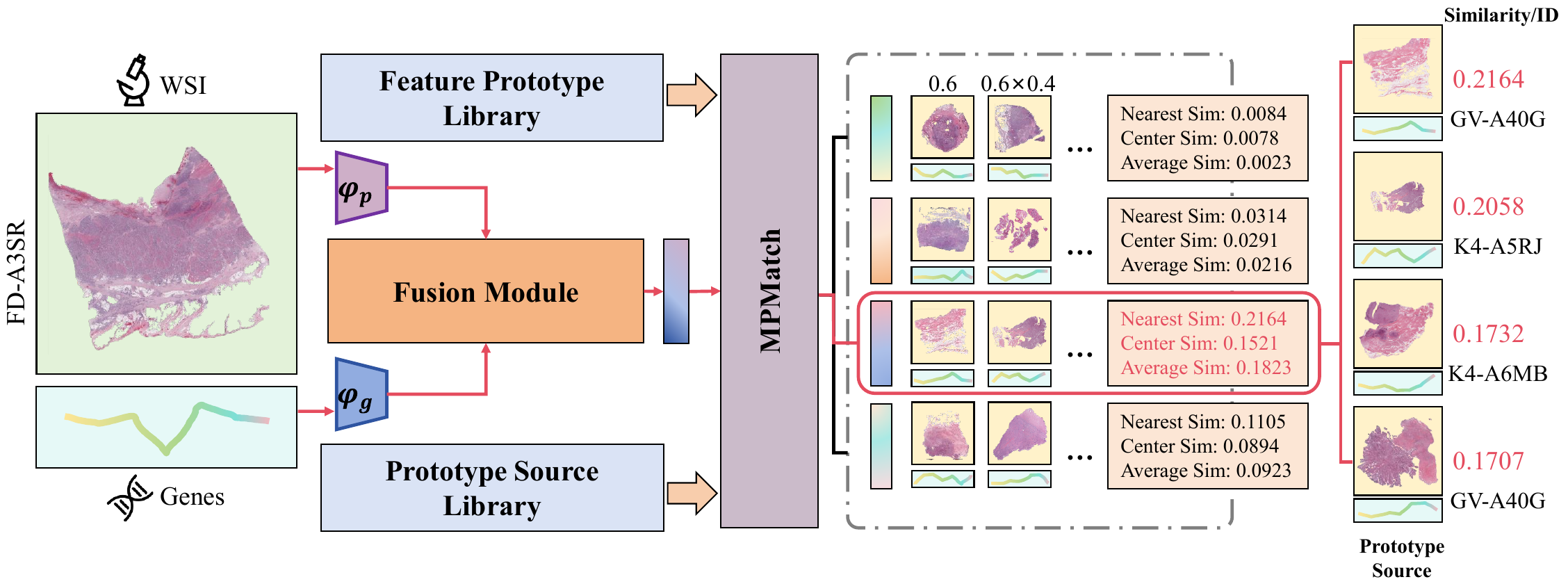} 
    \caption{Traceable interpretability analysis of FeatProto. Paired WSI and genomic inputs are fused into multimodal features and matched with the Feature Prototype Library by MPMatch using nearest, average, and center similarities. The matched prototypes are traced to source samples for interpretable risk prediction.}
    \label{fig: explain}
\end{figure*}

\subsection{Hyperparameter Sensitivity Analysis}

We report sensitivity analyses of several key hyperparameters under the same cross-validation protocol, with the purpose of examining model robustness to parameter choices. As demonstrated in Table \ref{table: simi}, PMDSim achieves the best performance by enhancing feature contrast via power scaling and reducing the impact of outliers, outperforming both Euclidean and cosine similarity. Table \ref{table: m_of_sim} presents the hyperparameter analysis of the power exponent $m$ in PMDSim, and results show that performance is optimal when \(m = 2\).

In the multi-level similarity structure of MPMatch, better matching performance is achieved with larger weights on $\alpha_{\mathrm{sim}}$ and $\beta_{\mathrm{sim}}$. In contrast, overreliance on the central prototype term $\gamma_{\mathrm{sim}}$ may overlook fine-grained prototype distributions and lead to mismatches, as reflected by the performance drop in Table~\ref{table: set}. The ratio $0.4\!:\!0.4\!:\!0.2$ achieves the best performance.
As shown in Table \ref{table: lambda}, for the update parameter in EMA ProtoUp, a small \(\lambda\) leads to an overly high proportion of new prototypes, thus neglecting the stabilizing effect of historical prototypes; conversely, a large \(\lambda\) results in too weak an influence of new prototypes, causing slow prototype library updates.

\subsection{Ablation Study}

The ablation study systematically assessed the contributions of the Wandering Prototype, EMA ProtoUp, MPMatch, and ProtoSurv Loss to the survival analysis performance on the LUAD and BLCA datasets. This was achieved by comparing the full model with variants that were missing each component. Tables \ref{table: loss_1} and \ref{table: ablation} confirm that all four modules enhance performance.

\subsubsection{Impact of the Wandering Prototype} The Wandering Prototype provides storage space for edge/exceptional samples, and pays extra attention to them during matching, in order to enhance the model's ability to generalize across different manifestations of the same diagnosis. Removing the Wandering Prototype led to a decrease in the LUAD C-index from 71.67\% to 69.03\% when using the CMTA base. In the BLCA dataset, the C-index dropped from 67.55\% to 67.03\% with the CMTA base and from 69.18\% to 68.89\% with the CCL base, underscoring its role in defining class boundaries.

\subsubsection{Role of EMA ProtoUp} EMA ProtoUp enables smooth updates to feature prototypes, avoiding instability caused by sudden changes in the feature space. Removing EMA ProtoUp resulted in a 3.69\% decrease in the LUAD C-index and a 1.47\% decrease in the BLCA C-index (CMTA base), indicating the importance of EMA ProtoUp for maintaining stability and accuracy. 

\subsubsection{Importance of MPMatch} MPMatch is an essential basis for prototype decision-making. Switching the MPMatch to a nearest matching approach caused a 4.97\% decrease in the LUAD C-index (CMTA base), which had the most significant impact on overall performance, emphasizing the value of the multi-level matching strategy in reducing interference from atypical samples.

\subsubsection{Performance of ProtoSurv Loss} As shown in Table~\ref{table: loss_1}, replacing ProtoSurv Loss with NLLSurvLoss or ProtoLoss leads to clear performance drops. On LUAD, the C-index decreases by 3.88\% and 7.96\%, respectively, and a similar trend is observed on BLCA. These results indicate that combining survival supervision with prototype optimization yields the best performance, whereas ProtoLoss alone performs the worst.

\subsection{Interpretability Analysis}

Fig. \ref{fig: explain} illustrates the traceable interpretability analysis of our method in the context of survival prediction inference and decision-making. The input sample, along with WSI bags and gene sequences, and the corresponding features are obtained through the Fusion Module of the backbone. In MPMatch, the Feature Prototype Library is retrieved for comparative analysis. Three similarity metrics are used to characterize the similarities with prototype libraries of different classes from both macro and micro perspectives. After obtaining the prediction results, tracing is conducted via the Prototype Source Library, which serves as the basis for interpretability analysis. We can further derive predictive diagnostic conclusions through specific comparisons between source samples and the input sample.

\section{Discussion}

To improve the interpretability and discriminative power of survival prediction, we propose FeatProto, which integrates global-local WSI features and genomic data into a unified prototype space. By introducing a Prototype Source Library together with EMA ProtoUp and MPMatch, FeatProto enables traceable prototype reasoning, dynamic prototype evolution, and multilevel similarity-based decision-making under tumor heterogeneity.

The traceable interpretability of FeatProto permeates the entire prediction process, distinguishing it from conventional post-hoc explanation methods such as Grad-CAM~\cite{Selvaraju_2017_ICCV} and SHAP~\cite{9265985}. In contrast to prior prototype-based multimodal survival frameworks such as MMP~\cite{song2024multimodal} and PIBD~\cite{zhang2023prototypical}, FeatProto additionally incorporates explicit prototype-to-source tracing and adaptive prototype updating. The prototype records and their mappings to raw data enable more direct localization of matched biological evidence, while the weighted fusion of three similarity levels in MPMatch provides a more quantitative basis for prediction. Clinically, such traceable multimodal evidence may help clinicians better verify whether a predicted risk is supported by plausible histopathological and genomic patterns. This may further improve the transparency of prognosis assessment and patient stratification for follow-up and treatment planning.

Nevertheless, this study still has several limitations. First, the number of prototypes is manually specified, and more adaptive allocation strategies may further improve flexibility. Second, although each instantiated model uses only a single pathology encoder, future work may explore multi-encoder fusion to better capture fine-grained histopathological structures. The additional results with CONCH suggest that FeatProto has promising generalizability across different pathology encoders, indicating that the proposed prototype-based reasoning mechanism is not restricted to a specific backbone. Third, strict external cross-center multimodal validation remains limited because public datasets with matched WSI, genomic profiles, and survival outcomes are still scarce. In addition, differences in sequencing types, gene panels, feature dimensions, and preprocessing pipelines hinder cross-cohort genomic comparison. Future work will focus on standardized cross-center multimodal cohorts and more robust cross-dataset alignment to support clinical translation.

\section{Conclusion}
We introduce FeatProto, an innovative framework designed to improve the interpretability of cancer survival predictions by using multimodal feature prototype learning. FeatProto leverages global-local WSI and genomic information within a dynamically updated prototype space to deliver superior predictive accuracy and clinical interpretability, offering an effective framework for multimodal prognostic modeling. It utilizes the EMA ProtoUp strategy to maintain stable cross-modal associations and incorporates the MPMatch module for robust inference. This framework captures both holistic and specific features of WSIs and genomic profiles, facilitating interpretable decision-making. Our research confirms that FeatProto's approach to prototype space reasoning delivers accurate and understandable prognostic results, making it ideally suited for clinical applications. Looking forward, we plan to adapt FeatProto for personalized oncology treatment and real-time prognostic monitoring, expanding its applicability and impact in the medical field.

\small
\bibliography{ref}
% \end{CJK}
\end{document}